\documentclass[sigconf]{acmart}

\AtBeginDocument{%
  \providecommand\BibTeX{{%
    \normalfont B\kern-0.5em{\scshape i\kern-0.25em b}\kern-0.8em\TeX}}}

\copyrightyear{2022}
\acmYear{2022}
\setcopyright{acmcopyright}
\acmConference[MM '22] {Proceedings of the 30th ACM International Conference on Multimedia }{October 10--14, 2022}{Lisbon, Portugal.}
\acmBooktitle{Proceedings of the 30th ACM International Conference on Multimedia (MM '22), October 10--14, 2022, Lisbon, Portugal}
\acmPrice{15.00}

\acmISBN{978-1-4503-9203-7/22/10}
\acmDOI{10.1145/3503161.3547986}

\usepackage{multirow}

\acmSubmissionID{989}

\begin{document}

\title{Learning Parallax Transformer Network for Stereo Image JPEG Artifacts Removal}


\author{Xuhao Jiang}
\orcid{0000-0002-4646-5052}
\affiliation{%
	\institution{School of Computer Science, Shanghai Key Laboratory of
		Intelligent Information Processing, Shanghai Collaborative Innovation Center of Intelligent Visual Computing, Fudan University}
	   \city{Shanghai}
	   \country{China}
}
\email{20110240011@fudan.edu.cn}

\author{Weimin Tan}
\orcid{0000-0001-7677-4772}
\affiliation{%
	\institution{School of Computer Science, Shanghai Key Laboratory of
		Intelligent Information Processing, Shanghai Collaborative Innovation Center of Intelligent Visual Computing, Fudan University}
	\city{Shanghai}
	\country{China}
}
\email{wmtan@fudan.edu.cn}

\author{Ri Cheng}
\orcid{0000-0002-5866-6847}
\affiliation{%
	\institution{School of Computer Science, Shanghai Key Laboratory of
		Intelligent Information Processing, Shanghai Collaborative Innovation Center of Intelligent Visual Computing, Fudan University}
	\city{Shanghai}
	\country{China}
}
\email{rcheng20@fudan.edu.cn}

\author{Shili Zhou}
\orcid{0000-0001-7283-2314}
\affiliation{%
	\institution{School of Computer Science, Shanghai Key Laboratory of
		Intelligent Information Processing, Shanghai Collaborative Innovation Center of Intelligent Visual Computing, Fudan University}
	   \city{Shanghai}
	   \country{China}
}
\email{slzhou19@fudan.edu.cn}

\author{Bo Yan}
\orcid{0000-0003-0256-9682}
\authornote{Corresponding Author. This work is supported by NSFC (Grant No.: U2001209, 61902076) and Natural Science Foundation of Shanghai (21ZR1406600).}
\affiliation{%
	\institution{ School of Computer Science, Shanghai Key Laboratory of
		Intelligent Information Processing, Shanghai Collaborative Innovation Center of Intelligent Visual Computing, Fudan University}
	\city{Shanghai}
	\country{China}
}
\email{byan@fudan.edu.cn}

\renewcommand{\shortauthors}{Xuhao Jiang et al.}


\begin{abstract}
Under stereo settings, the performance of image JPEG artifacts removal can be further improved by exploiting the additional information provided by a second view. However, incorporating this information for stereo image JPEG artifacts removal is a huge challenge, since the existing compression artifacts make pixel-level view alignment difficult. In this paper, we propose a novel parallax transformer network (PTNet) to integrate the information from stereo image pairs for stereo image JPEG artifacts removal. Specifically, a well-designed symmetric bi-directional parallax transformer module is proposed to match features with similar textures between different views instead of pixel-level view alignment. Due to the issues of occlusions and boundaries, a confidence-based cross-view fusion module is proposed to achieve better feature fusion for both views, where the cross-view features are weighted with confidence maps. Especially, we adopt a coarse-to-fine design for the cross-view interaction, leading to better performance. Comprehensive experimental results demonstrate that our PTNet can effectively remove compression artifacts and achieves superior performance than other testing state-of-the-art methods.

\end{abstract}




\iffalse


\fi



\maketitle

\section{Introduction}

\begin{figure}[t]
	\begin{center}
		\includegraphics[width=1\linewidth]{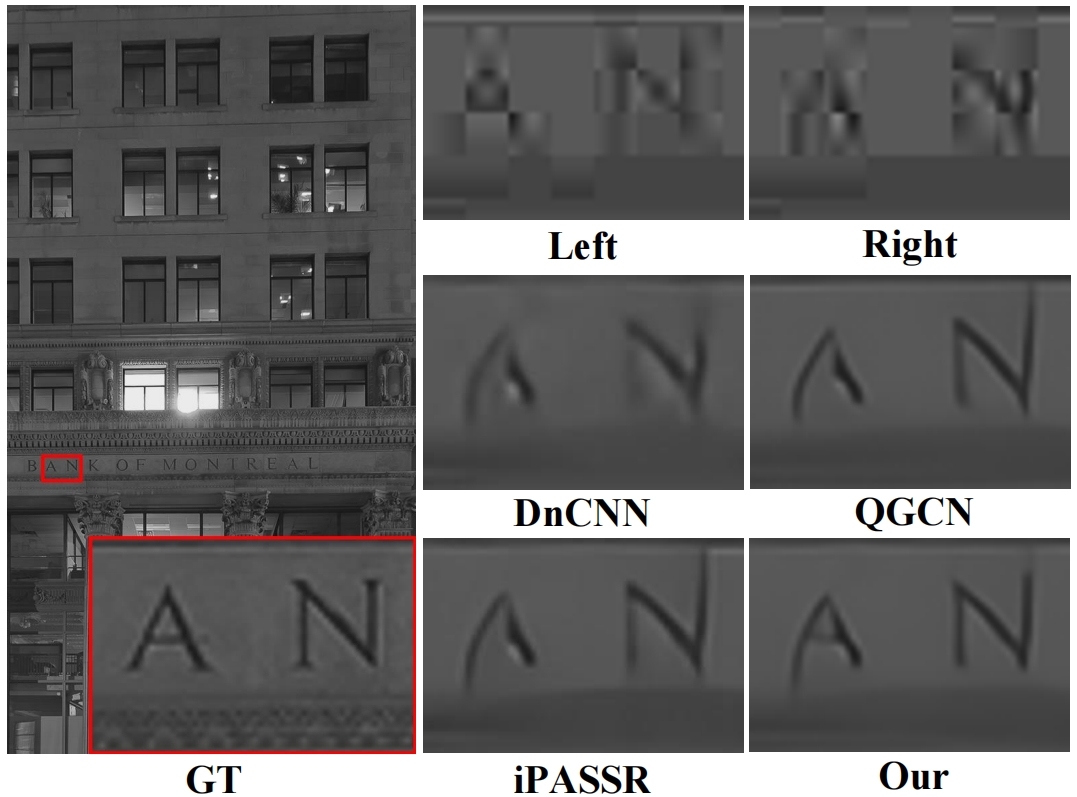}
	\end{center}
	\caption{ The deblocking results on image '0007' from Flickr1024~\cite{wang2019flickr1024} at quality factor 10. On the left is the ground-truth (GT). The first row on the right is the corresponding left and right compressed image patches. Compared to the state-of-the-art methods (DnCNN~\cite{zhang2017beyond}, QGCN~\cite{li2020learning} and iPASSR~\cite{wang2021symmetric}), our PTNet can generate better results due to the effective use of information from both views. }
	\label{fig:shouye}
\end{figure}
With recent advances in dual cameras, stereo images have shown great commercial value in many practical applications, including smartphones and autonomous vehicles. Usually, stereo images require a large number of bits to store the information from two views, resulting in the challenges of storage and transmission. Image compression algorithms can help to reduce the data size of the original digital stereo images, but inevitably introduce complex compression noise, such as blocking artifacts~\cite{dong2015compression}. This may lead to the degradation of the visual quality and the performance of the subsequent vision tasks. Therefore, exploring methods for compressed stereo image artifacts removal is urgently needed, especially for the widely used JPEG format.

JPEG is one of the most widely used image compression algorithms, and its processing procedure consists of four steps, including block division, discrete cosine transformation (DCT), quantization and entropy coding. The block-based JPEG compression algorithm ignores spatial correlations between image blocks, which results in image discontinuities at block boundaries. To cope with this problem, the early approaches~\cite{chen2001adaptive,foi2007pointwise,liu2018graph} focus on filter design or employ various optimizations, but still suffer from blurring the images. Some deep learning-based approaches~\cite{dong2015compression,fu2019jpeg,zhang2017beyond,zhang2019residual,jiang2021towards} with novel architectures attempt to remove the compression artifacts by learning a nonlinear mapping between the compressed and original images. These above methods are all designed for single image deblocking, and to the best of our knowledge, no work has been conducted on stereo image deblocking. While these algorithms can also be used to recover the left and right images independently, their performance may be severely limited due to the lack of additional information from another view. Especially, some details are lost in one view, but may exist in another view (as shown in Fig.~\ref{fig:shouye}).

Recently, some methods~\cite{jeon2018enhancing,wang2019learning,song2020stereoscopic,wang2021symmetric} have been proposed for stereo image super-resolution, which is the most relevant research topic for the stereo image deblocking task. Wang \emph{et al.}~\cite{wang2019learning} design a parallax-attention module to handle different stereo images with large disparity variations. The well-designed parallax-attention module utilizes the predicted transformation matrix to achieve view alignment. Some follow-up methods~\cite{song2020stereoscopic,wang2021symmetric} improve the stereo correspondence performance by improving the parallax attention module. Although these methods achieve good results in stereo image super-resolution, they perform poorly in stereo deblocking. The main reason is that the compression artifacts destroy the stereo correspondence between the two views, resulting in the difficulties of pixel-level alignment. Therefore, we consider using the transformer to perform a robust matching search on the reference view instead of pixel-level alignment.

\begin{figure*}[t]
	\begin{center}
		\includegraphics[width=1\linewidth]{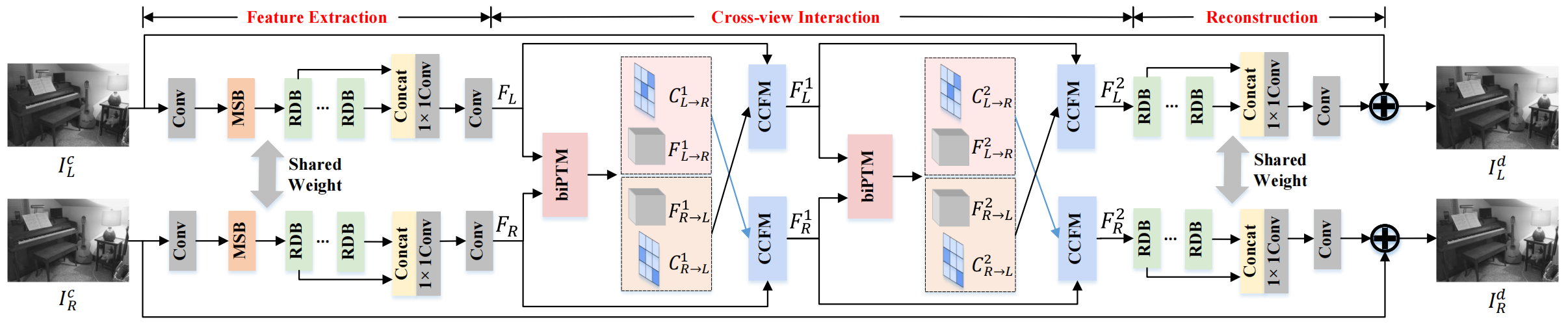}
	\end{center}
	\caption{ The architecture of the proposed PTNet. The proposed biPTM is designed to achieve cross-view interaction, which is based on the mutual attention mechanism of two different views. CCFM is designed to effectively fuse the cross-view features, which can help to solve the issues of occlusions and boundaries. In addition, MSB is a well-designed multi-scale feature extraction Module, and RDB is the residual dense block~\cite{zhang2018residual}. The details of MSB and RDB can be found in appendix.}
	\label{fig:framework}
\end{figure*}

In this paper, we propose a novel parallax transformer network (PTNet) to integrate the information from stereo image pairs for stereo image JPEG artifacts removal. The details of overall framework are shown in Fig.~\ref{fig:framework}. We design a symmetric bi-directional parallax transformer module (biPTM) to computes the relevance between left and right image features and further match these features, enabling cross-view interaction. Specifically, for any region in the target view, we use the mutual attention mechanism to extract the region features with the highest relevance in the reference view, and use them to enhance the target region. Note that the goal of biPTM is to find the required reference features for the target regions and does not do view alignment, so it can perform well even with significant disparities and compression artifacts. Considering the issue of occlusion, a confidence-based cross-view fusion module (CCFM) is proposed to effectively integrate cross-view information. To achieve better cross-view interaction, we adopt a coarse-to-fine design that utilizes the enhanced features for further cross-view feature matching. To sum up, our main contributions are as follows:

\begin{itemize}
	\item We propose a novel parallax transformer network for stereo image JPEG artifacts removal, which exploits the information complementarity between left and right compressed images to achieve better stereo image deblocking. To the best of our knowledge, this is the first effort to address this task.

	\item  A novel symmetric bi-directional parallax transformer module is proposed to implement cross-view interaction, which is based on the mutual attention mechanism and achieves effective feature matching.

	\item Considering the occlusion issues, we propose a confidence-based cross-view fusion module that enables effective feature fusion for both views.
		
	\item Our approach achieves the state-of-the-art performance as compared to recent single-image JPEG artifacts removal methods and a stereo image super-resolution method.

\end{itemize}

\section{Related Work}
\subsection{JPEG Artifacts Removal}

JPEG artifacts removal has been studied for a long time and notable progress has been achieved in the past few years. Early methods~\cite{zhang2013compression,chen2001adaptive,foi2007pointwise} attempt to remove the compression artifacts by designing specific filters. Others treat JPEG artifacts removal as an ill-posed inverse problem and solve it by using sparse representation~\cite{chang2013reducing}, graph~\cite{liu2018graph} and regression trees~\cite{jancsary2012loss}.

Witnessing the recent success of convolutional neural networks (CNNs) in most computer vision tasks~\cite{he2016deep,long2015fully}, learning-based methods~\cite{dong2015compression,fu2019jpeg,zhang2017beyond,zhang2019residual,li2020learning,wang2020jpeg,Fu2021TNNS,zhen2020CSVT,galteri2019deep,jiang2021towards,9607618,zhang2020residual} have recently attracted a lot of attention, and have been explored for image deblocking. Zhang~\emph{et al.}~\cite{zhang2017beyond} utilize the residual learning~\cite{he2016deep} and batch normalization~\cite{ioffe2015batch} to speed up the training process as well as boost the deblocking performance. Fu~\emph{et al.}~\cite{fu2019jpeg} design a deep convolutional sparse coding (DCSC) network architecture to effectively reduce JPEG artifacts by using dilated convolution~\cite{yu2015multi}. QGCN~\cite{li2020learning} is able to handle a wide range of quality factors due to the novel utilization of the quantization tables as part of the training data. 

However, the existing methods are all designed for single image deblocking, and their performance is limited in stereo deblocking since additional information from another view is not exploited. In this paper, we propose a novel parallax transformer network which exploits the information complementarity between two views to achieve better stereo image deblocking.
                        
\subsection{Stereo Image Super-Resolution}

In recent years, many deep learning-based methods~\cite{jeon2018enhancing,wang2019learning,song2020stereoscopic,wang2021symmetric,ying2020stereo} have been proposed to tackle the problem of stereo image super-resolution, and achieve promising results. Wang~\emph{et al.}~\cite{wang2019learning} try to combine stereo matching and stereo image super-resolution, and propose a parallax attention network named as PASSRnet, which can cope with the issue of varying parallax. Especially, the proposed parallax-attention network can capture stereo correspondence. Inspired by~\cite{wang2019learning}, Song~\emph{et al.}~\cite{song2020stereoscopic} propose a self and parallax attention network to aggregate the information from its own view and the second view simultaneously. On the basis of PASSRnet, Wang~\emph{et al.}~\cite{wang2021symmetric} make a symmetrical design and propose iPASSR, which can super-resolve both sides of views within a single inference. These parallax attention-based methods all attempt to capture the stereo correspondence and warp the features of the second view to the target view at the pixel level, thereby improving the super-resolution performance of the target view.

However, the above methods are not suitable for stereo image deblocking task and show poor performance. The main reason is that the compression artifacts destroy the original texture information of the image, which makes pixel-level view alignment difficult. As shown in Fig.~\ref{fig:motivation}, the matching regions also show different textures after being compressed. Unlike these methods, our method attempts to find the most relevant features in both views, which is achieved by a robust transformer-based matching. In particular, for the occlusions and boundaries, we can also find the most relevant matching features for them, and use a confidence-based weighting method for feature fusion.

\subsection{Vision Transformer}
Recently, Transformer-based models~\cite{khan2021transformers,han2020survey,yang2020learning} have achieved promising performance in various vision tasks, such as image
recognition~\cite{dosovitskiy2020image,touvron2021training}, object detection~\cite{carion2020end,zhu2020deformable} and video understanding~\cite{girdhar2019video}. Some approaches are designed for image restoration~\cite{chen2021pre,9607618,wang2021uformer}. Chen~\emph{et al.}~\cite{chen2021pre} study the low-level computer vision task (e.g., denoising, super-resolution and deraining) and develop a new pre-trained model. These methods focus on the feature fusion based on self-attention mechanism, and aim to achieve excellent performance. However, unlike previous methods, we design a symmetric bi-directional parallax transformer module to achieve prediction of parallax information, which is then used for stereo image feature matching. Especially, the proposed module builds a mutual attention mechanism between information from two views, and performs stereo image feature matching.

\section{Method}

\subsection{Motivation}
\begin{figure}[t]
	\begin{center}
		\includegraphics[width=1\linewidth]{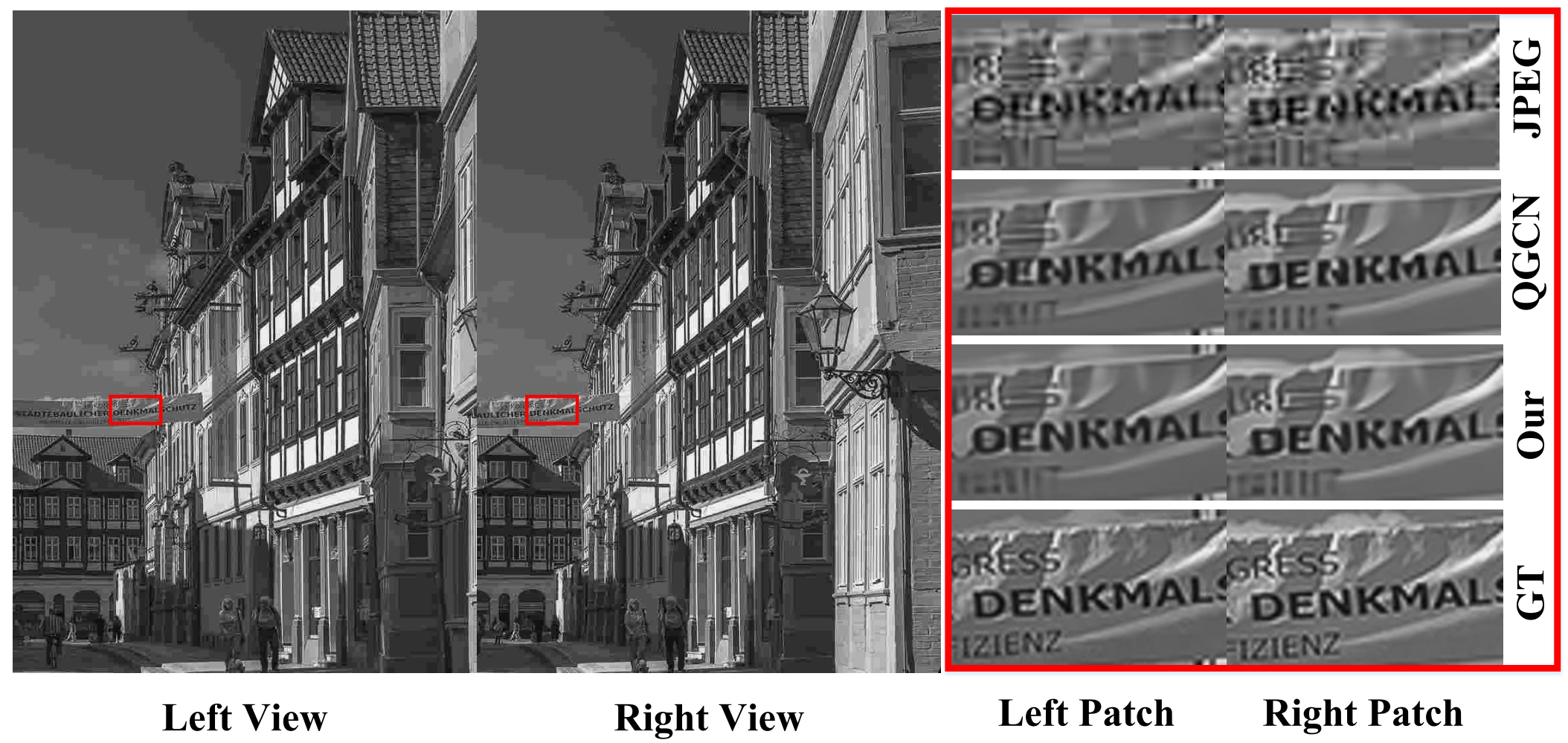}
	\end{center}
	\caption{ An example of stereo image deblocking results on image ’0001’ from
		Flickr1024~\cite{wang2019flickr1024} at quality factor 10. On the left is the JPEG-compressed stereo image pair. Note that we mark the matching regions of the left and right views with a red box. On the right are the marked regions, their deblocking results of QGCN~\cite{li2020learning} and our PTNet, and the corresponding ground-truth.  }
	\label{fig:motivation}
\end{figure}

There are numerous matching regions in the left and right views of a stereo image pair. When the stereo image pair is compressed, these regions are significantly degraded and usually exhibit similar degraded contents. As shown in Fig.~\ref{fig:motivation}, on the left is a JPEG-compressed stereo image pair. In this stereo image pair, most of the regions match each other despite occlusions and boundaries. For in-depth analysis, a matching region is selected as an example and marked in the images. On the right of Fig.~\ref{fig:motivation}, we provide a zoomed-in view of the region (first row) and its corresponding ground-truth (GT) (fourth row). We find that although the GT patches of the left and right view are similar, their corresponding compressed patches have different details. Specifically, the letter N in the left patch is clearer while the letter A in the right patch is clearer. This inspires us to attempt using the information of two views simultaneously for stereo image deblocking, since the information of the two views can complement each other.

There are two main reasons for this phenomenon: 1) Existence of parallax between two views. This causes the matching regions of two views to be similar but not completely consistent. 2) Block-based compression processing. The JPEG compression algorithm uses $8\times8$ blocks as the basic processing unit, which may cause overlaps. For example, a letter falls in two processing units simultaneously in one view, but only exists in one unit in another view. These two reasons cause the matching regions to show different degradations when they are JPEG compressed. Therefore, the information of two views are complementary. Benefiting from the binocular information, our results may achieve better results than single-image deblocking algorithms, as shown in Fig.~\ref{fig:motivation}.

\subsection{Overview of Our PTNet}

The goal of our PTNet is to reconstruct the deblocking results ($I^{d}_{L}$, $I^{d}_{R}$) from a JPEG-compressed stereo image pair ($I^{c}_{L}$, $I^{c}_{R}$), aiming to keep deblocking results ($I^{d}_{L}$, $I^{d}_{R}$) and the corresponding uncompressed stereo image pair ($I_{L}$, $I_{R}$) consistent in pixel. The architecture of our PTNet is shown in the Fig.~\ref{fig:framework}, which mainly consists of three parts: feature extraction, cross-view interaction and reconstruction. Note that the entire network is symmetric and the weights of its left and right branches are shared.

Specifically, given ($I^{c}_{L}$, $I^{c}_{R}$), we first extract the features ($F_{L}$, $F_{R}$) of the left and right images separately, which are used for subsequent feature matching and reconstruction. This process is denoted as,
\begin{equation}
	F_L=H_{FE}(I^c_L),~F_R=H_{FE}(I^c_R),
\end{equation}
where $H_{FE}(\cdot)$ represents the feature extraction module. Following the previous works~\cite{   fu2019jpeg,wang2020jpeg}, we design a multi-scale feature extraction block (MSB) to enhance the feature extraction capability of the model. In addition, we also adopt four residual dense blocks (RDBs)~\cite{zhang2018residual} in our model. The details of MSB and RDB can be found in appendix.

These extracted features ($F_L$, $F_R$) are then used for feature matching and feature enhancement in the cross-view interaction module. This module adopts the coarse-to-fine design and is mainly divided into two stages. Each stage consists of one bi-directional parallax transformer module (biPTM) and one confidence-based cross-view fusion module (CCFM). In the first stage, we achieve effective cross-view information interaction, and we further enhance the information interaction in the second stage. Especially, since the first stage utilizes the binocular information to enhance the features of two views, the second stage can achieve more accurate feature matching. This can be expressed as,
\begin{equation}
	F^1_L,F^1_R=H_{CVI^1}(F_L,~F_R),~F^2_L,F^2_R=H_{CVI^2}(F^1_L,~F^1_R),
\end{equation}
where $H_{CVI^1}(\cdot)$ and $H_{CVI^2}(\cdot)$ stand for the functions of two stages in the cross-view interaction module respectively. The details of biPTM and CCFM will be explained in later sections.

Finally, these features ($F^2_L$, $F^2_R$) are used in the reconstruction module to generate our deblocking results. This module is mainly composed of four RDBs. Aiming to reconstruct better results, we also add a global residual design. This can be expressed as
\begin{equation}
	I^{d}_{L}, I^{d}_{R}=H_{R}(F^2_L,~F^2_R,~I^{c}_{L},~I^{c}_{R}),
\end{equation} 
where $H_{R}(\cdot)$ represents the reconstruction module.

\subsection{Bi-Directional Parallax Transformer}

\begin{figure}[t]
	\begin{center}
		\includegraphics[width=1\linewidth]{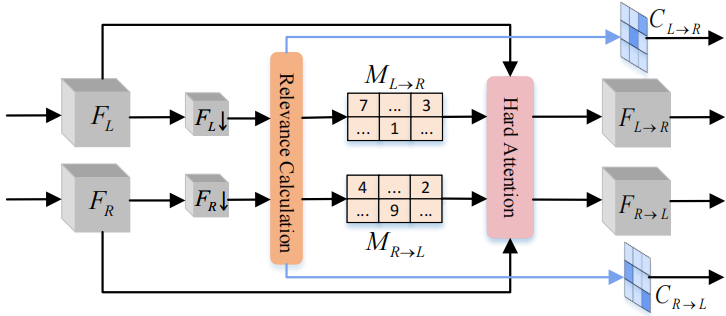}
	\end{center}
	\caption{Architecture of the proposed bi-directional parallax transformer module (biPTM). $F_L$ and $F_R$ represent the feature maps of left view and right view. $F_L\downarrow$ and $F_R\downarrow$ are obtained by downsampling $F_L$ and $F_R$. $M_{L\rightarrow R}$ and $M_{R\rightarrow L}$ indicate the hard attention maps, which are computed from relevance calculation module and used to match feature maps of different views. $F_{L\rightarrow R}$ and $F_{R\rightarrow L}$ are the converted features. $C_{L\rightarrow R}$ and $C_{R\rightarrow L}$ are the corresponding confidence maps.}
	\label{fig:transformer}
\end{figure}

The compression artifacts cause difficulties in pixel-level view alignment, and inaccurate alignment may affect the performance of stereo image deblocking. Therefore, we consider finding the required reference features for the target region without view alignment. We utilize the mutual attention mechanism to match features with similar textures between different views. To this end, a symmetric bi-directional parallax transformer module (biPTM) is proposed, which is shown in Fig.~\ref{fig:transformer}. Our biPTM takes the features of the left and right view as input, and outputs the cross-view converted features and their confidence maps. Note that the cross-view conversion of our two-view features is symmetric. Here, we introduce the calculation process of the feature conversion from the left view to the right view in detail.

Firstly, the left and right image features ($F_L$, $F_R$) are downsampled by a factor of 4, which can effectively reduce the calculation amount of the module. We make the three basic elements of the attention mechanism inside a transformer as
\begin{equation}
	Q=F_R\downarrow,~K=F_L\downarrow,~V=F_L,
\end{equation}
where Q, K, V represent query, key and value respectively. Q and K are unfolded into patches and normalized, denoted as
\begin{equation}
	\bar{q}_i=\frac{q_i}{||q_i||}~(i\in[1,H_{F_R\downarrow}\times W_{F_R\downarrow}]),
\end{equation} 
\begin{equation}
    \bar{k}_j=\frac{k_j}{||k_j||}~(j\in[1,H_{F_L\downarrow}\times W_{F_L\downarrow}]),
\end{equation} 
where $H_{F_R}$ and $W_{F_R}$ represent height and width of $F_R$, $H_{F_L}$ and $W_{F_L}$ represent height and width of $F_L$, respectively. Then we calculate the relevance R between the left and right features ($F_L$, $F_R$) by estimating the similarity between Q and K in the relevance calculation module. This can be expressed as,
\begin{equation}
	R=Q\cdot K^T
\end{equation} 
where R consists of $i\times j$ probability values $r_{ij}$.

After that, we use a hard attention mechanism to weight $V$ for each query $q_i$ based on $R$. Therefore, only the most relevant features in $V$ are converted for each query $q_i$ by using the hard attention mechanism. The hard attention map $M_{L\rightarrow R}$ can be obtained by finding the maximum probability of $R$ in the $j$ dimension. This can be expressed as,
\begin{equation}
	m_i = \underset{j}{\arg \max} r_{ij},~c_i = \underset{j}{\max} r_{ij},
\end{equation} 
where the value of $m_i$ in $M_{L\rightarrow R}$ is a coordinate index, which means the most relevant position in $F_L$ corresponds to the $i^{th}$ position in $F_R$, the value of $c_i$ is the probability value of $m_i$. Then we unfold the $V$ into patches, and each patch is four times the size of $q_i$, denoted as $v_i~(i\in[1,H_{F_R\downarrow}\times W_{F_R\downarrow}])$. Based on the obtained $M_{L\rightarrow R}$, an index selection operation is used to process $v_i$ to obtain the converted patch $z_i$, denoted as $z_i = v_{m_i}$.

Finally, the converted patch $z_i$ is folded to generate the converted features $F_{L\rightarrow R}$. Since the matching probability value of occlusions and boundaries will be relatively low, the probability value $c_i$ can be used to generate the confidence map $C_{L\rightarrow R}$ by using folding operation. Similarly, we can obtain $F_{R\rightarrow L}$ and $C_{R\rightarrow L}$ by resetting Q, K and V as,
\begin{equation}
	Q=F_L\downarrow,~K=F_R\downarrow,~V=F_R.
\end{equation} 
To simplify the calculation, we obtain the corresponding relevance by transposing the previously obtained $R$.

\subsection{Cross-View Feature Fusion}

\begin{figure}[t]
	\begin{center}
		\includegraphics[width=1\linewidth]{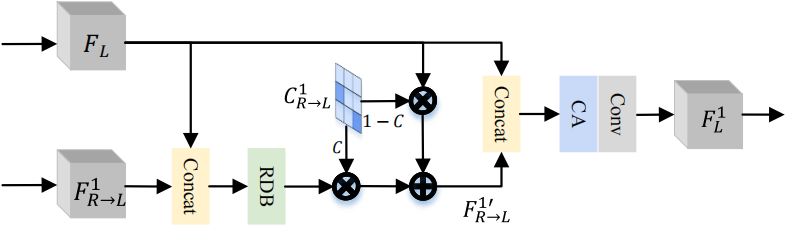}
	\end{center}
	\caption{ Architecture of the proposed confidence-based cross-view fusion module (CCFM). $F_L$ and $F_{R\rightarrow L}$ represent the feature maps of left view and the converted feature maps of right view. $C_{R_L}$ stands for the confidence map of $F_{R_L}$. $F^{'}_{L}$ represents the fused features. RDB is the residual dense block~\cite{zhang2018residual}, and CA~\cite{hu2018squeeze} represents the channel attention module.}
	\label{fig:fusion}
\end{figure}

Due to the issues of occlusions and boundaries in stereo image processing, these occlusion and boundary regions do not match well with another view. To address this problem, we propose a confidence-based cross-view fusion module (CCFM) to achieve effective cross-view feature fusion, in which the cross-view features are weighted with the confidence maps produced by biPTM. The details of CCFM are shown in Fig.~\ref{fig:fusion}. Note that the weights of CCFM are shared, and the corresponding calculation process is symmetric in the left and right branches.

Here, we introduce the fusion process of $F_L$ and $F^1_{R\rightarrow L}$ in detail. First, $F_L$ is concatenated with $F^1_{R\rightarrow L}$ and fed into one RDB~\cite{zhang2018residual} for initial feature fusion. We consider that regions with high confidence are more inclined to adopt converted features $F^1_{R\rightarrow L}$, and regions with low confidence adopt the features of the target view $F_L$. Therefore, a confidence-based weighting method is designed to fuse $F_L$ and $F^1_{R\rightarrow L}$. This can be expressed as,
\begin{equation}
	F^{1'}_{R\rightarrow L}=C^1_{R\rightarrow L}\odot f_{RDB}([F_L, F^1_{R\rightarrow L}])+(1-C^1_{R\rightarrow L})\odot F_L
\end{equation} 
where $f_{RDB}$ represents the function of RDB. With the help of this confidence-based weighting method, occluded regions of converted features $F^1_{R\rightarrow L}$ can be filled with
the corresponding features $F_L$ from the target view, leading to continuous spatial distributions. Finally, $F^{1'}_{R\rightarrow L}$ is concatenated with $F_L$ again, and then fed to a channel attention layer (CA)~\cite{hu2018squeeze} and a convolution layer to generate the final fused features $F^1_L$. Similarly, we can obtain $F^1_R$, $F^2_L$ and $F^2_R$ by following the same calculation process with different input features.

\subsection{Optimization }
Given a training dataset with $N$ stereo image pairs~$\{I^i_L,~I^i_R\}^N_{i=1}$, we can obtain its corresponding JPEG-compressed stereo image pairs~$\{I^{c,i}_L,~I^{c,i}_R\}^N_{i=1}$ and the reconstructed results~$\{I^{d,i}_L,~I^{d,i}_R\}^N_{i=1}$. Following the previous works~\cite{li2020learning,fu2019jpeg}, we also adopt the $l_1$~norm for network training, since $l_1$~norm can yield the sharper image results. The loss function is denoted as,
\begin{equation}
	L=\frac{1}{N}\sum_{i=1}^{N}\{||I^i_L-I^{d,i}_L||_1+||I^i_R-I^{d,i}_R||_1\}.
\end{equation} 

During our PTNet training, Pytorch is used as the training toolbox, and the Adam optimization algorithm [50] with $\beta1 = 0.9$, $\beta2 = 0.999$, and a mini-batch size of 48 is adopted. All the experiments are conducted on three NVIDIA GeForce RTX 1080 Ti. The learning rate is changed from $2\times10^{-4}$ to $2\times10^{-6}$ at the interval of twenty epochs. The training was stopped after 60 epochs since more epochs do not provide further consistent improvement.

\begin{table*}[t!]
	\small
	\caption{Performance comparisons of various methods based on the grayscale left images from Flickr1024~\cite{wang2019flickr1024}, KITTI2012~\cite{geiger2012we}, KITTI2015~\cite{menze2015object} and Middlebury~\cite{scharstein2014high}. Here, PSNR|SSIM|PSNR-B values achieved on the left images (\emph{i.e., Left}) are reported. The best results are boldfaced.
	}\label{tab:left}% ???? ????
	\begin{tabular}{c|c|c|c|c|c|c|c}
		\hline
		Dataset                     & QF   & JPEG               & DnCNN~\cite{zhang2017beyond}              & DCSC~\cite{fu2019jpeg}               & QGCN~\cite{li2020learning}               & iPASSR~\cite{wang2021symmetric}             & \textbf{PTNet} \\ \hline
		\multirow{3}{*}{Flickr1024} & 10             & 25.99/0.7868/23.72 & 27.40/0.8231/27.02 & 27.56/0.8287/27.15 &27.72/0.8351/27.43  &27.76/0.8342/27.21  &  \textbf{28.05/0.8403/27.54}        \\ \cline{2-8} 
		                            & 20             & 28.08/0.8614/25.75 & 29.66/0.8895/29.05 & 29.84/0.8926/29.16 &30.09/0.8975/29.51  &30.12/0.8973/29.42  &  \textbf{30.39/0.9017/29.59}        \\ \cline{2-8} 
		                            & 30             & 29.42/0.8938/27.14 & 31.09/0.9172/30.39 & 31.26/0.9190/30.48 &31.58/0.9243/30.85  &31.58/0.9232/30.77  &  \textbf{31.83/0.9264/30.89}        \\ \hline
		\multirow{3}{*}{KITTI2012}  & 10             & 29.27/0.8292/26.46 & 30.82/0.8665/30.53 & 30.99/0.8711/30.65 &31.20/0.8759/30.95  &31.01/0.8716/30.55  &  \textbf{31.43/0.8786/31.05}        \\ \cline{2-8} 
		                            & 20             & 31.72/0.8919/28.89 & 33.28/0.9152/32.78 & 33.42/0.9175/32.92 &33.60/0.9201/33.26  &33.46/0.9186/33.04  &  \textbf{33.85/0.9231/33.30}        \\ \cline{2-8} 
		                            & 30             & 33.07/0.9170/30.27 & 34.65/0.9347/34.02 & 34.80/0.9362/34.18 &34.97/0.9388/34.46  &34.85/0.9372/34.30  &  \textbf{35.18/0.9404/34.48}        \\ \hline
		\multirow{3}{*}{KITTI2015}  & 10             & 29.31/0.8230/26.22 & 30.90/0.8615/30.53 & 31.06/0.8665/30.60 &31.31/0.8714/\textbf{30.96}  &31.05/0.8669/30.48  &  \textbf{31.42/0.8730}/30.92        \\ \cline{2-8} 
		                            & 20             & 32.02/0.8937/28.75 & 33.59/0.9177/32.88 & 33.72/0.9200/33.00 &33.96/0.9226/\textbf{33.27}  &33.77/0.9211/33.15  &  \textbf{34.07/0.9245/33.27}        \\ \cline{2-8} 
		                            & 30             & 33.54/0.9220/30.23 & 35.13/0.9401/34.20 & 35.26/0.9415/34.39 &35.46/0.9436/\textbf{34.63}  &35.32/0.9424/34.58  &  \textbf{35.57/0.9449/}34.58        \\ \hline
		\multirow{3}{*}{Middlebury} & 10             & 29.65/0.8114/27.09 & 31.38/0.8529/31.22 & 31.57/0.8582/31.38 &31.85/0.8643/31.73  &31.67/0.8602/31.38  &  \textbf{32.05/0.8676/31.88}       \\ \cline{2-8} 
		                            & 20             & 32.06/0.8826/29.43 & 33.79/0.9081/33.42 & 33.98/0.9111/33.64 &34.26/0.9156/34.03  &34.12/0.9136/33.84  &  \textbf{34.51/0.9200/34.12}       \\ \cline{2-8} 
		                            & 30             & 33.40/0.9110/30.86 & 35.16/0.9304/34.70 & 35.35/0.9325/34.95 &35.54/0.9361/35.23  &35.46/0.9349/35.14  &  \textbf{35.85/0.9400/35.40}        \\ \hline
	\end{tabular}
\end{table*}

\begin{table*}[t!]
	\small
	\caption{Performance comparisons of various methods based on the grayscale stereo image pairs from Flickr1024~\cite{wang2019flickr1024}, KITTI2012~\cite{geiger2012we}, KITTI2015~\cite{menze2015object} and Middlebury~\cite{scharstein2014high}. Here, PSNR|SSIM|PSNR-B values achieved on the stereo image pairs (\emph{i.e., (Left + Right) /2}) are reported. The best results are boldfaced.
	}\label{tab:both}% ???? ????
	\begin{tabular}{c|c|c|c|c|c|c|c}
		\hline
		Dataset                     & QF & JPEG              & DnCNN~\cite{zhang2017beyond}              & DCSC~\cite{fu2019jpeg}              & QGCN~\cite{li2020learning}               & iPASSR~\cite{wang2021symmetric}              & \textbf{PTNet} \\ \hline
		\multirow{3}{*}{Flickr1024}& 10             & 26.00/0.7860/23.74 & 27.41/0.8223/27.03 & 27.57/0.8279/27.16 &27.74/0.8345/27.44  &27.78/0.8335/27.22  &  \textbf{28.07/0.8397/27.55}                   \\ \cline{2-8}
		                           & 20             & 28.09/0.8607/25.76 & 29.67/0.8889/29.06 & 29.85/0.8920/29.17 &30.10/0.8970/29.53  &30.13/0.8967/29.43  &  \textbf{30.41/0.9011/29.61}                   \\ \cline{2-8}
		                           & 30             & 29.43/0.8933/27.15 & 31.09/0.9166/30.40 & 31.26/0.9185/30.49 &31.59/0.9240/30.86  &31.58/0.9227/30.77  &  \textbf{31.83/0.9259/30.90}                   \\ \hline
		\multirow{3}{*}{KITTI2012} & 10             & 29.12/0.8267/26.33 & 30.64/0.8641/30.33 & 30.81/0.8687/30.44 &31.00/0.8732/30.75  &30.83/0.8693/30.35  &  \textbf{31.23/0.8761/30.83}                  \\ \cline{2-8} 
		                           & 20             & 31.52/0.8897/28.71 & 33.05/0.9131/32.51 & 33.19/0.9154/32.65 &33.36/0.9180/32.99  &33.24/0.9166/32.78  &  \textbf{33.61/0.9209/33.01}                   \\ \cline{2-8} 
		                           & 30             & 32.85/0.9149/30.08 & 34.40/0.9327/33.72 & 34.55/0.9343/33.89 &34.71/0.9374/\textbf{34.18}  &34.61/0.9353/34.02  &  \textbf{34.92/0.9384/34.18}                  \\ \hline
		\multirow{3}{*}{KITTI2015} & 10             & 29.72/0.8314/26.57 & 31.37/0.8708/31.04 & 31.54/0.8740/31.12 &31.82/0.8807/31.50  &31.53/0.8760/31.00  &  \textbf{31.97/0.8831/31.52}                  \\ \cline{2-8} 
		                           & 20             & 32.55/0.9008/29.20 & 34.16/0.9245/33.54 & 34.30/0.9268/33.67 &34.57/0.9292/34.01  &34.35/0.9278/33.81  &  \textbf{34.73/0.9319/34.02}                  \\ \cline{2-8} 
		                           & 30             & 34.13/0.9279/30.73 & 35.76/0.9455/34.93 & 35.90/0.9469/35.12 &36.13/0.9490/\textbf{35.46}  &35.96/0.9478/35.30  &  \textbf{36.28/0.9507/}35.39                   \\ \hline
		\multirow{3}{*}{Middlebury}& 10             & 29.62/0.8105/27.02 & 31.32/0.8518/31.14 & 31.53/0.8572/31.25 &31.74/0.8624/31.48  &31.62/0.8594/31.26  &  \textbf{32.03/0.8672/31.75}                  \\ \cline{2-8} 
		                           & 20             & 32.03/0.8827/29.35 & 33.76/0.9084/33.30 & 33.96/0.9113/33.48 &34.22/0.9164/33.71  &34.10/0.9140/33.69  &  \textbf{34.51/0.9207/33.97}                   \\ \cline{2-8} 
		                           & 30             & 33.38/0.9112/30.76 & 35.15/0.9310/34.57 & 35.35/0.9331/34.79 &35.57/0.9368/35.07  &35.48/0.9356/35.01  &  \textbf{35.88/0.9409/35.25}                    \\ \hline
	\end{tabular}
\end{table*}

\section{Experiments}

\subsection{Datasets and Evaluation}
Following iPASSR~\cite{wang2021symmetric}, we also use 60 images from Middlebury~\cite{scharstein2014high} and 800 images from Flickr1024~\cite{wang2019flickr1024} as the training dataset. For test, we adopt 5 images from Middlebury, 20 images from KITTI 2012~\cite{geiger2012we}, 20 images from KITTI 2015~\cite{menze2015object}, and 112 images from Flickr1024 as the test dataset, which is the same as iPASSR. To train the proposed PTNet, the images are first cropped into patches of size $64\times 160$ with a stride of 20. These patches are then processed by JPEG compression algorithm with a random quality factor $QF\in[10,30]$ to get the corresponding compressed image patches. In this paper, Python Image Library (PIL) is adopted to encode images into JPEG format, since it employs a standard quantization table proposed by the Independent JPEG Group. In addition, these patches are randomly flipped horizontally and vertically
for data augmentation. We only focus on the restoration of the luminance channel (in YCrCb space) in this paper.

Following~\cite{fu2019jpeg,li2020learning}, we apply the PSNR, structural similarity (SSIM)~\cite{wang2004image}, and PSNR-B~\cite{yim2010quality} to evaluate the model performance. Referring to iPASSR~\cite{wang2021symmetric}, we report PSNR, SSIM and PSNR-B scores on the left view (\emph{i.e., Left}) and the average PSNR, SSIM and PSNR-B scores on stereo image pairs (\emph{i.e., (Left + Right) /2}).

\subsection{Comparison against SOTA Methods}
In this section, the proposed PTNet and the state-of-the-art
algorithms including DnCNN~\cite{zhang2017beyond}, DCSC~\cite{fu2019jpeg}, QGCN~\cite{li2020learning} and iPASSR~\cite{wang2021symmetric} are compared quantitatively and qualitatively. DnCNN, DCSC and QGCN are single image deblocking methods, and iPASSR is a high-performance stereo image super-resolution method. To conduct a fair comparison, DnCNN and QGCN are finetuned on the training dataset for 10 epochs. We use the pre-trained model of DCSC to test its performance due to the unavailability of the training code. For iPASSR, we set its scale factor to 1, and use the luminance channel as input. Then iPASSR can be trained on the training dataset for stereo image deblocking.

\textbf{Quantitative results.} Tables \ref{tab:left} and \ref{tab:both} show the quantitative results on four datasets with JPEG QF 10, 20 and 30, respectively. Specifically, Table \ref{tab:left} shows the performances of all test algorithm on the left view. It can be found that the proposed PTNet achieves the best performance at all JPEG QF. Compared with the single image deblocking methods, our PTNet achieves a significant performance improvement. The main reason is that PTNet makes full use of the information of two views and achieves better deblocking results. Although iPASSR also takes information of two views as input, it does not take into account that compression artifacts destroy stereo correspondence, and inaccurate feature warping leads to poor performance. In contrast, our PTNet still performs well in the presence of compression artifacts. To comprehensively evaluate the performance of stereo image deblocking, we report the average performance on two views, and the experimental results in Table \ref{tab:both} also confirm that our PTNet outperforms other compared methods.

\begin{figure*}[t]
	\begin{center}
		\includegraphics[width=0.95\linewidth]{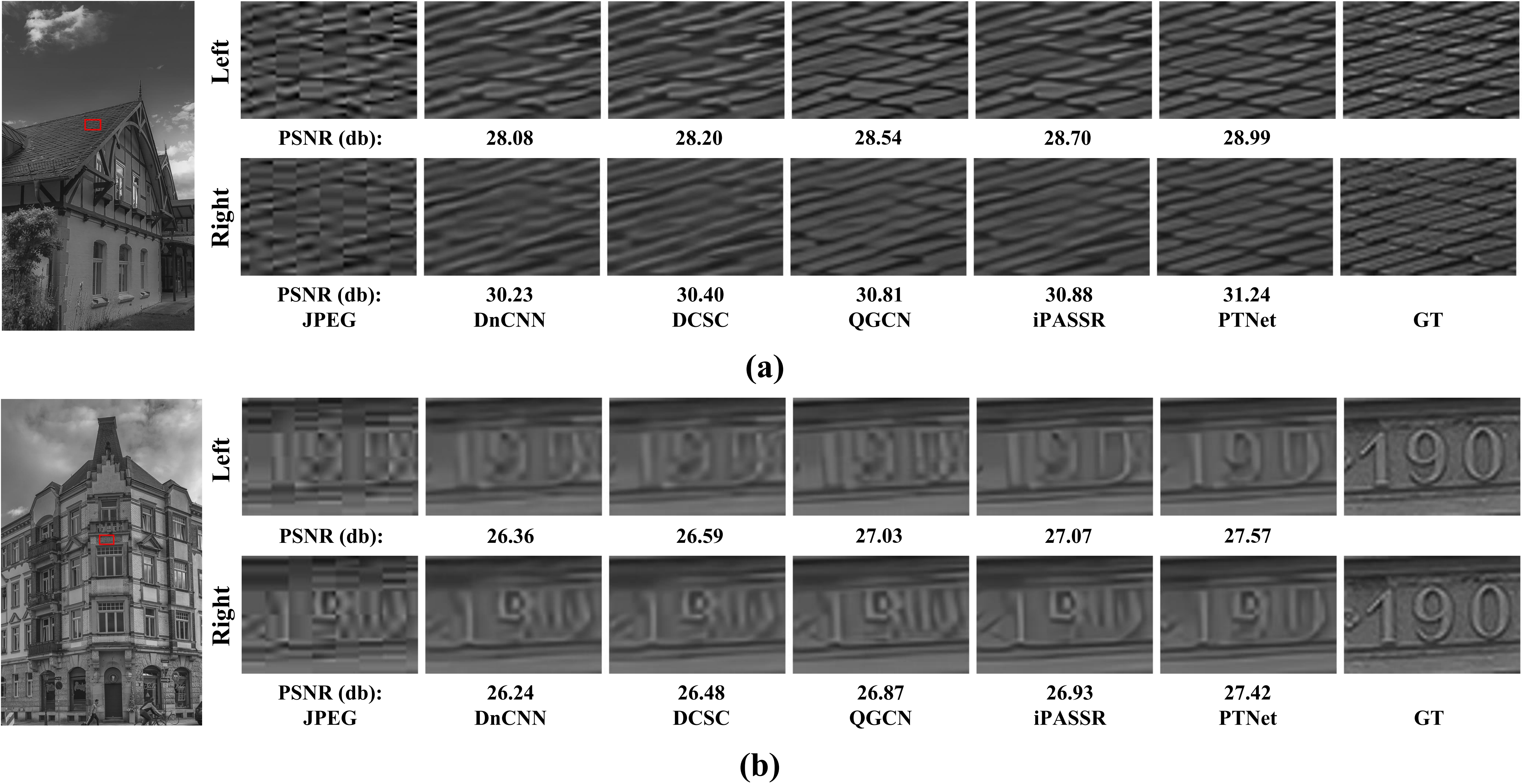}
	\end{center}
	\caption{ Visual comparisons on the images '0003' (a) and '0043' (b) from Flickr1024~\cite{wang2019learning} at QF 10. The proposed PTNet is compared with the state-of-the-art methods including DnCNN~\cite{zhang2017beyond}, DCSC~\cite{fu2019jpeg}, QGCN~\cite{li2020learning} and iPASSR~\cite{wang2021symmetric}. The first row shows the deblocking results on the left view, while the second row shows the deblocking results on the right view. The number below each image patch represents the PSNR value. Note that our PTNet can produce better results compared to other methods.}
	\label{fig:visual_1}
\end{figure*}

\begin{table*}[t!]
	\small
	\caption{Performance comparisons between variations of our PTNet based on the grayscale images from Flickr1024~\cite{wang2019flickr1024}, KITTI2012~\cite{geiger2012we}, KITTI2015~\cite{menze2015object} and Middlebury~\cite{scharstein2014high}. Here, PSNR|SSIM|PSNR-B values achieved on the left images (\emph{i.e., Left}) are reported. The best results are boldfaced.
	}\label{tab:ablation}% ???? ????
	\begin{tabular}{c|c|c|c|c|c}
		\hline
		Dataset                     & QF             & w/o biPTM \& CCFM              & w/o CCFM               & w/o CTF             & \textbf{PTNet} \\ \hline
		\multirow{3}{*}{Flickr1024} & 10             &27.85/0.8347/27.34 &27.98/0.8395/27.49  &28.01/0.8392/27.54  & \textbf{28.05/0.8403/27.54}        \\ \cline{2-6} 
		& 20             &30.17/0.8975/29.39 &30.32/0.9009/29.55  &30.35/0.9011/29.58  & \textbf{30.39/0.9017/29.59}        \\ \cline{2-6} 
		& 30             &31.62/0.9233/30.68 &31.76/0.9259/30.85  &31.78/0.9258/30.89  & \textbf{31.83/0.9264/30.89}        \\ \hline
		\multirow{3}{*}{KITTI2012}  & 10             &31.14/0.8735/30.79 &31.36/0.8778/31.02  &31.39/0.8781/ \textbf{31.05}  & \textbf{31.43/0.8786/31.05}        \\ \cline{2-6} 
		& 20             &33.57/0.9196/33.04 &33.76/0.9225/33.26  &33.82/0.9228/33.29  & \textbf{33.85/0.9231/33.30}        \\ \cline{2-6} 
		& 30             &34.93/0.9379/34.23 &35.11/0.9401/34.45  &35.13/0.9402/34.43  & \textbf{35.18/0.9404/34.48}        \\ \hline
		\multirow{3}{*}{KITTI2015}  & 10             &31.19/0.8687/30.74 &31.38/0.8726/30.90  &31.39/0.8724/30.90  & \textbf{31.42/0.8730/30.92}        \\ \cline{2-6} 
		& 20             &33.85/0.9218/33.11 &34.00/0.9241/33.24  &34.03/0.9241/33.24  & \textbf{34.07/0.9245/33.27}        \\ \cline{2-6} 
		& 30             &35.39/0.9430/34.44 &35.50/0.9445/34.57  &35.52/0.9444/34.54  & \textbf{35.57/0.9449/34.58}        \\ \hline
		\multirow{3}{*}{Middlebury} & 10             &31.77/0.8614/31.59 &31.99/0.8669/31.81  &32.00/0.8666/31.82  & \textbf{32.05/0.8676/31.88}       \\ \cline{2-6} 
		& 20             &34.19/0.9143/33.81 &34.45/0.9193/\textbf{34.12}  &34.45/0.9193/34.08  & \textbf{34.51/0.9200/34.12}       \\ \cline{2-6} 
		& 30             &35.51/0.9351/35.06 &35.79/0.9394/35.39  &35.80/0.9394/35.39  & \textbf{35.85/0.9400/35.40}        \\ \hline
		\hline
		\emph{Params.}		&-&0.90 M&0.90 M&0.91 M&0.91 M					\\ \hline
	\end{tabular}
\end{table*}

\textbf{Qualitative results.} The proposed PTNet can produce deblocking results with high perceptual quality, and the qualitative comparison results are shown in Fig.\ref{fig:visual_1}. Compared to other methods, our PTNet can remove compression artifacts more effectively and recover high-fidelity textures. The main reason is that PTNet makes good use of the additional information provided by the second view. Although iPASSR also utilizes information from two views for stereo image deblocking, its reconstructed results are more blurry than ours, because inaccurate pixel-level stereo matching may affect the performance of feature fusion.

\subsection{Ablation Study}

\begin{figure*}[t]
	\begin{center}
		\includegraphics[width=0.95\linewidth]{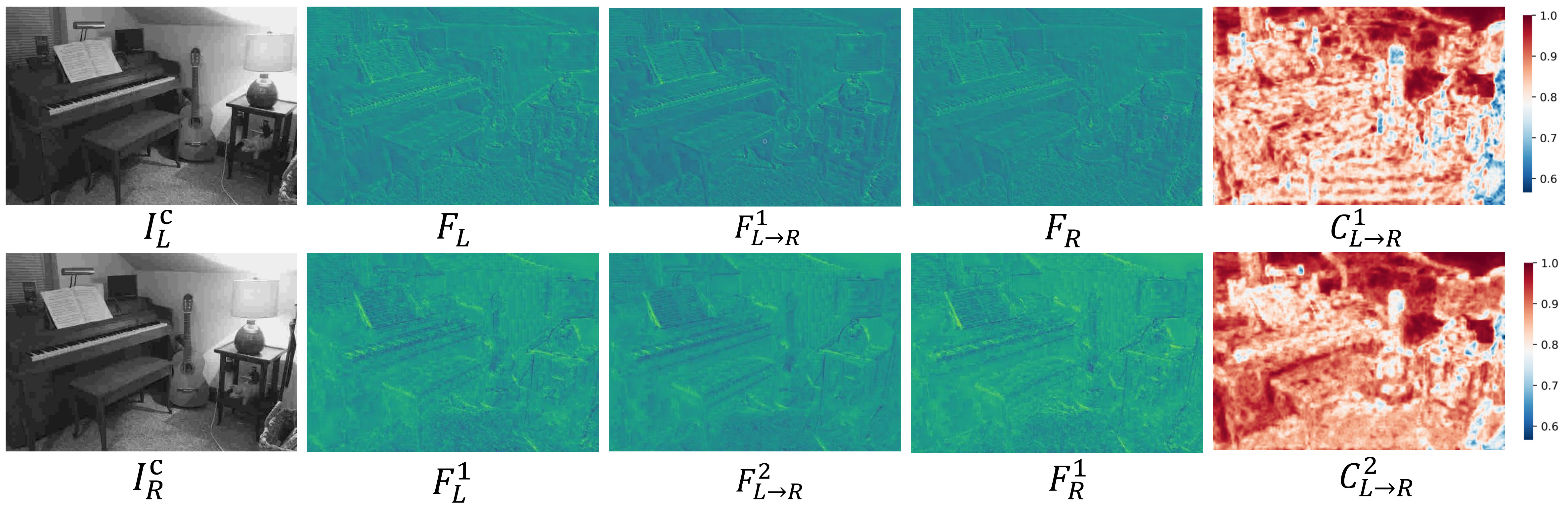}
	\end{center}
	\caption{ Visualization of feature maps generated by our PTNet on the image 'piano' from Middlebury~\cite{scharstein2014high}. Since PTNet is symmetric, we only show feature matching from the left view to the right view. The first column is the compressed images at QF 10. The first row shows the feature maps of the first stage in the cross-view interaction, including $F_L$, $F^1_{L\rightarrow R}$, $F_R$ and $C^1_{L\rightarrow R}$. The second row shows the feature maps of the second stage, including $F^1_L$, $F^2_{L\rightarrow R}$, $F^1_R$ and $C^2_{L\rightarrow R}$. Better zoom in.}
	\label{fig:visualization}
\end{figure*}

\begin{table}[t!]
	\small
	\caption{Performance comparisons between iPASSR and iPASSR+.
	}\label{tab:iPASSR}% ???? ????
	\begin{tabular}{c|c|c|c}
		\hline
		Dataset                     & QF & iPASSR & iPASSR+              \\ \hline
		\multirow{3}{*}{Flickr1024} & 10    &27.76/0.8342/27.21         & 27.92/0.8361/27.48         \\ \cline{2-4} 
		& 20    &30.12/0.8973/29.42         & 30.30/0.8998/29.57          \\ \cline{2-4} 
		& 30    &31.58/0.9232/30.77         & 31.74/0.9250/30.85         \\ \hline
		\multirow{3}{*}{KITTI2012}  & 10    &31.01/0.8716/30.55         & 31.26/0.8751/30.93                   \\ \cline{2-4} 
		& 20    &33.46/0.9186/33.04         & 33.74/0.9214/33.25                   \\ \cline{2-4} 
		& 30    &34.85/0.9372/34.30         & 35.07/0.9393/34.42                    \\ \hline
		\multirow{3}{*}{KITTI2015}  & 10    &31.05/0.8669/30.48         & 31.30/0.8697/30.85                   \\ \cline{2-4} 
		& 20    &33.77/0.9211/33.15         & 33.98/0.9230/33.24                    \\ \cline{2-4} 
		& 30    &35.32/0.9424/34.58         & 35.48/0.9437/34.52                     \\ \hline
		\multirow{3}{*}{Middlebury} & 10    &31.67/0.8602/31.38         & 31.92/0.8641/31.75                      \\ \cline{2-4} 
		& 20    &34.12/0.9136/33.84         & 34.42/0.9182/34.10                     \\ \cline{2-4} 
		& 30    &35.46/0.9349/35.14         & 35.76/0.9384/35.36                     \\ \hline
	\end{tabular}
\end{table}
In this section, we study and analyze the contributions of different modules to our PTNet, including the bi-directional parallax transformer module (biPTM), the confidence-based cross-view fusion module (CCFM) and the coarse-to-fine (CTF) structure. To this end, we remove these modules from our PTNet separately. Since confidence maps are not available when biPTM is removed, we remove both biPTM and CCFM to verify the effectiveness of biPTM. We also add several RDBs, and several convolutional layers in the variation of our PTNet, aiming to keep similar model size. We test the performances of PTNet without biPTM and CCFM (w/o biPTM \& CCFM), PTNET without CCFM (w/o CCFM) and PTNet without CTF (w/o CTF). Specifically, w/o biPTM \& CCFM concatenates the features of two views for fusion, w/o CCFM removes the operation of the feature weighting calculation and w/o CTF only uses one stage for cross-view interaction. The experimental results are shown in Table~\ref{tab:ablation}. It can be found that the performances of three variations all decrease compared with PTNet on all datasets. This confirms that our proposed modules can effectively improve the performance of the model for stereo image deblocking. Note that our PTNet has a significant performance improvement compared to w/o biPTM \& CCFM. This means that biPTM contributes the most to the improvement of model performance.

In addition, we also conduct a comparative experiment to further confirm that our biPTM can indeed improve the performance for stereo image deblocking. We replace the view alignment module in iPASSR with biPTM, and name this model iPASSR+. As shown in Table~\ref{tab:iPASSR}, the performance of iPASSR+ is significantly improved on all datasets. This demonstrates the effectiveness of our biPTM for stereo image deblocking.

\subsection{Visualization Results}

To more intuitively show that our biPTM can achieve good cross-view feature matching, we visualize the features of both stages of biPTM, as shown in Fig.~\ref{fig:visualization}.

Firstly, we can find that $F_L$ and $F_R$ are not aligned, and concatenating them for fusion does not achieve good cross-view interaction, which is confirmed by the ablation experiments. Our biPTM can provide an efficient converted feature $F^1_{L\rightarrow R}$ for $F_R$ even with significant artifacts in the images. Specifically, in the corresponding regions, $F^1_{L\rightarrow R}$ has the texture features that match the $F_R$, so better cross-view feature fusion can be achieved. In addition, we can make similar conclusions in the second stage through observation. 

Secondly, the confidence map $C^1_{L\rightarrow R}$ shows small confidence values at the boundaries, which is consistent with the observation of the input stereo image pair. Note that the unconfident regions of $C^2_{L\rightarrow R}$ become smaller in the second stage, which also verifies that the features enhanced by the first stage can achieve more reliable feature matching.

\section{Conclusion}

In this paper, we investigate the problem of stereo image JPEG artifacts removal for the first time and provide an in-depth analysis. To this end, we propose a novel parallax transformer network (PTNet) to simultaneously remove compression artifacts from two views. Specifically, we design a symmetric bi-directional parallax transformer module (biPTM) to computes the relevance between the features of two views, and further match these features, enabling cross-view interaction. Due to the issues of occlusions and boundaries, a confidence-based cross-view fusion module (CCFM) is proposed to effectively integrate cross-view information. Experimental results demonstrate that our PTNet outperforms the test SOTA methods, and extensive ablation studies are performed to verify the effectiveness of our proposed modules. Furthermore, the proposed method can also be feasibly extended to cope with other stereo image processing tasks, such as stereo image deblurring. In the future, we will further explore the possibility of our method for different stereo image processing tasks.


\bibliographystyle{ACMMM}
\bibliography{stereo_deblocking}

\appendix
\section{Appendix}
\subsection{The Architectures of MSB and RDB}
\begin{figure}[h]
	\begin{center}
		\includegraphics[width=0.95\linewidth]{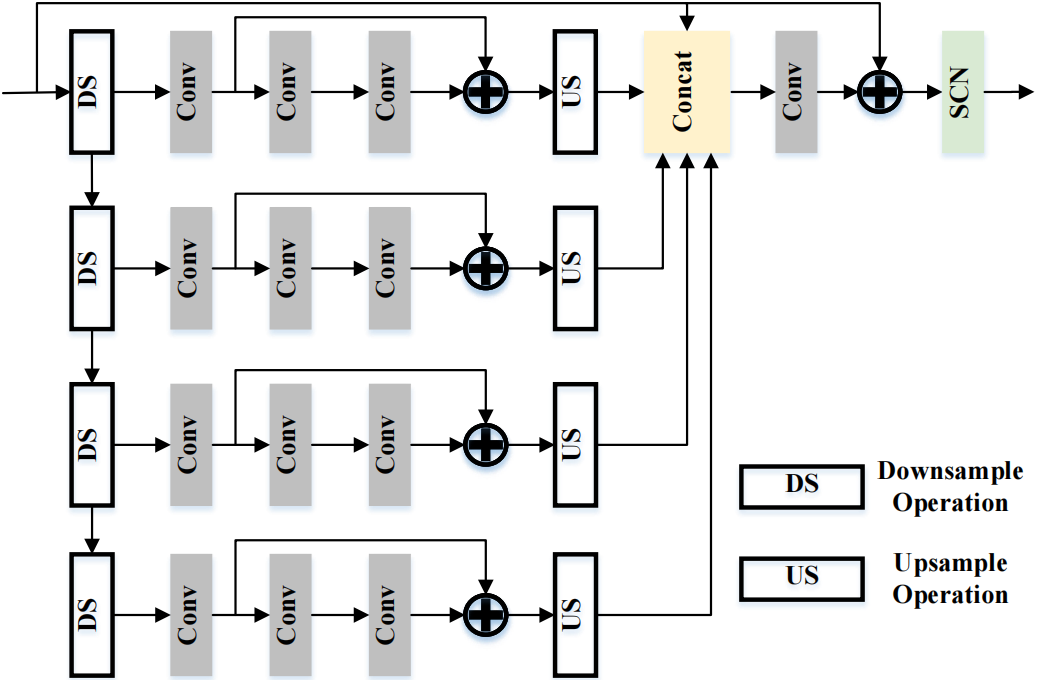}
	\end{center}
	\caption{Architecture of the multi-scale feature extraction Module. This module adopts the structure of downsampling-residual learning-upsampling to extract multi-scale features, then aggregate these features. Note that SCN~\cite{liu2020improving} is a self-calibrating module and can generate better feature representations.}
	\label{fig:MSB}
\end{figure}

The details of the multi-scale feature extraction Module (MSB) in the proposed PTNet are shown in Fig.~\ref{fig:MSB}. MSB adopts the structure of downsampling-
residual learning-upsampling to extract multi-scale features, then aggregate these features. Note that SCN~\cite{liu2020improving} is a self-calibrating module and can generate better feature representations. The size of the filter is $3\times3$ and the stride is 1. In addition, the channel number of the convolutional layers is 32.

\begin{figure}[h]
	\begin{center}
		\includegraphics[width=0.95\linewidth]{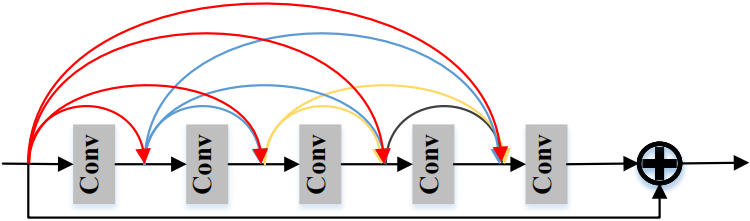}
	\end{center}
	\caption{Architecture of the residual dense block (RDB)~\cite{zhang2018residual}.}
	\label{fig:RDB}
\end{figure}

Witnessing the great success of the residual dense block (RDB)~\cite{zhang2018residual}, we also adopt RDB to extract features and reconstruct results. The details of RDB are shown in Fig.~\ref{fig:RDB}. RDB consists of five convolutional layers. The size of the filter is $3\times3$ and the stride is 1 with the channel number of 32.

\subsection{Model Size and Computational Complexity}
We compare the models in terms of the model size (Params: M), computational complexity (FLOPs: G) and inference efficiency (Runtime: s). Note that the FLOPs and runtime of the single image deblocking methods need to be doubled. Specifically, given a pair of stereo images with a resolution of 100$\times$100, then we calculate the FLOPs of each method. In addition, we calculate the average runtime required for a pair of stereo images on Middlebury  dataset. The experimental results are shown in Table~\ref{tab:model}. It can be found that our algorithm performs well in terms of model size, computational complexity and inference efficiency.

\begin{table}[h]
	\small
	\caption{Model size and computational complexity comparison of different methods.  
	}\label{tab:model}% ???? ????
	\begin{tabular}{c|c|c|c}
		\hline
		Method            & Params. (M)              & FLOPs (G)              & Runtime (s)               \\ \hline
		DnCNN              &0.67 &2$\times$6.66  &2$\times$0.129         \\ \cline{1-4}
		DCSC              &0.32 &2$\times$60.66  &2$\times$0.742         \\ \cline{1-4} 
		QGCN              &4.97 &2$\times$12.44  &2$\times$0.528        \\ \cline{1-4}  
		iPASSR              &1.43 & 27.16 & 0.754       \\ \cline{1-4}
		\textbf{PTNet}              &\textbf{0.91} & \textbf{16.64} &\textbf{0.578}        \\ \cline{1-4}    
		\hline
	\end{tabular}
\end{table}

\subsection{Hard and Soft Attention Mechanism}

The purpose of using hard attention mechanism is to improve the computational efficiency of the proposed biPTM. Compared with the soft attention mechanism, the hard attention mechanism has a faster computation speed. During conducting experiments, we also find that using the hard attention mechanism can improve the performance of the model at different QFs. We report the results on Middlebury dataset in Table~\ref{tab:attention}. Here, (PSNR, SSIM, PSNRB) are used to evaluate the performance. We can find that using the hard attention mechanism leads to better results than using the soft attention mechanism. Since the left and right views are highly similar, using a hard attention mechanism to match the most relevant features can achieve effective cross-view information interaction.

\begin{table}[h]
	\small
	\caption{Performance comparisons of different attention mechanism on Middlebury dataset. Here, PSNR|SSIM|PSNR-B values achieved on the left images (i.e., Left) are reported.  
	}\label{tab:attention}% ???? ????
	\begin{tabular}{c|c|c}
		\hline
		QF           & Soft Attention              & \textbf{Hard Attention} \\ \hline
		 10              &31.82/0.8631/31.70 &\textbf{32.05/0.8676/31.88}          \\ \cline{1-3}
		 20             &34.29/0.9159/33.94 &\textbf{34.51/0.9200/34.12}           \\ \cline{1-3} 
		 30               &35.60/0.9367/35.29 &\textbf{35.85/0.9400/35.40}          \\ \cline{1-3}    
		\hline
	\end{tabular}
\end{table}

\subsection{Disparity Estimation}

Following~\cite{wang2021symmetric}, we also test the performances of stereo matching at all test QFs. We obtain the stereo image deblocking results on the SceneFlow dataset~\cite{mayer2016large} by using different methods. Then, we utilize GwcNet~\cite{guo2019group} to evaluate the disparity. End-point-error (EPE) and t-pixel error rate (> tpx) are utilized as quantitative metrics to evaluate the predicted disparity. The experimental results are shown in Table~\ref{tab:disparity-evaluation}. Compared with other methods, the stereo matching performance of our results has been greatly improved, which confirms that our method is beneficial to disparity estimation.

\begin{table*}[h!]
	\small
	\caption{Quantitative results achieved by GwcNet~\cite{guo2019group} at QF 10. All these metrics are averaged on the test set of the SceneFlow dataset~\cite{mayer2016large}, where lower values indicate better performance. 
	}\label{tab:disparity-evaluation}% ???? ????
	\begin{tabular}{c|c|c|c|c|c|c|c}
		\hline
		Metric                     & QF   & JPEG               & DnCNN~\cite{zhang2017beyond}              & DCSC~\cite{fu2019jpeg}               & QGCN~\cite{li2020learning}               & iPASSR~\cite{wang2021symmetric}             & \textbf{PTNet} \\ \hline
		\multirow{3}{*}{EPE} & 10             &6.67  &6.35  &6.43  &5.64  &5.82  & \textbf{4.20}        \\ \cline{2-8} 
		& 20            &4.01  &3.34  &3.46  &3.82  &3.29  &\textbf{2.63}       \\ \cline{2-8} 
		& 30            &3.12  &2.51  &2.51  &2.86  &2.49  &\textbf{2.14}        \\ \hline
		\multirow{3}{*}{>1px(\%)}  & 10    &56.8  &49.7  &48.9  &46.3  &46.9  &\textbf{39.9}        \\ \cline{2-8} 
			& 20             &40.8  &31.5  &32.0  &31.8  &29.6  &\textbf{26.0}        \\ \cline{2-8} 
			& 30             &30.9  &23.5  &24.0  &24.8  &22.4  &\textbf{20.4}       \\ \hline
			\multirow{3}{*}{>2px(\%)}  & 10           &36.3  &29.1  &28.5  &26.2  &26.9  &\textbf{21.3}        \\ \cline{2-8} 
				& 20             &22.8  &15.9  &16.3  &16.4  &15.1  &\textbf{12.9}        \\ \cline{2-8} 
				& 30             &16.3  &11.7  &12.0  &12.7  &11.4  &\textbf{10.3}       \\ \hline
				\multirow{3}{*}{>3px(\%)} & 10           &25.1  &19.8  &19.5  &18.2  &18.3  &\textbf{14.4}       \\ \cline{2-8} 
					&20             &14.8  &10.9  &11.1  &11.9  &10.5  &\textbf{8.9}      \\ \cline{2-8} 
					&30&10.9&8.3&8.3&9.2&8.3&\textbf{7.4}        \\ \hline
				\end{tabular}
			\end{table*}

\subsection{More Visualizations}
\begin{figure*}[t]
	\begin{center}
		\includegraphics[width=1\linewidth]{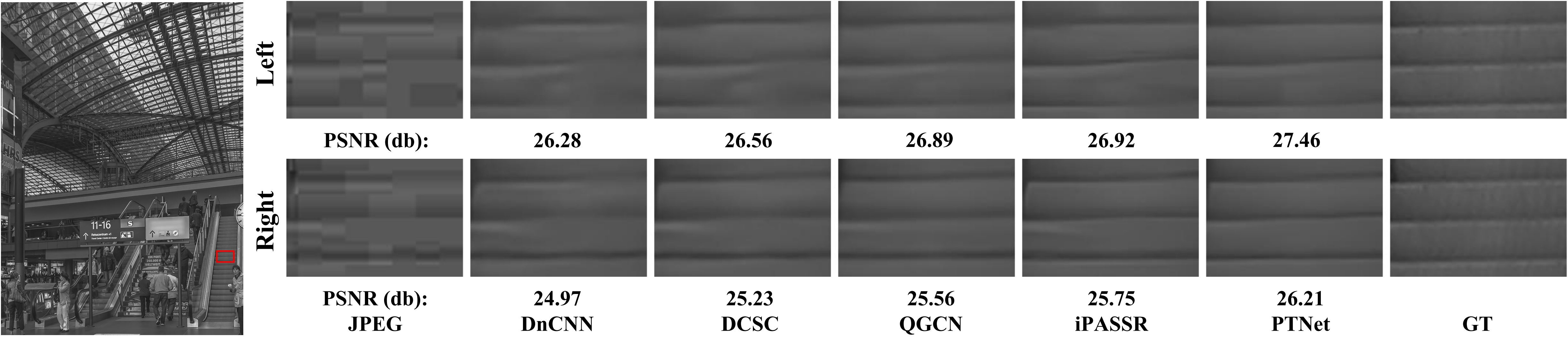}
	\end{center}
	\caption{ Visual comparisons on the image ’0035’ from Flickr1024~\cite{wang2019flickr1024} at QF 10. The proposed PTNet is compared with the state-of-the-art methods including DnCNN~\cite{zhang2017beyond}, DCSC~\cite{fu2019jpeg}, QGCN~\cite{li2020learning} and iPASSR~\cite{wang2021symmetric}. Better zoom in.}
	\label{fig:visual_re_1}
\end{figure*}

\begin{figure*}[t]
	\begin{center}
		\includegraphics[width=1\linewidth]{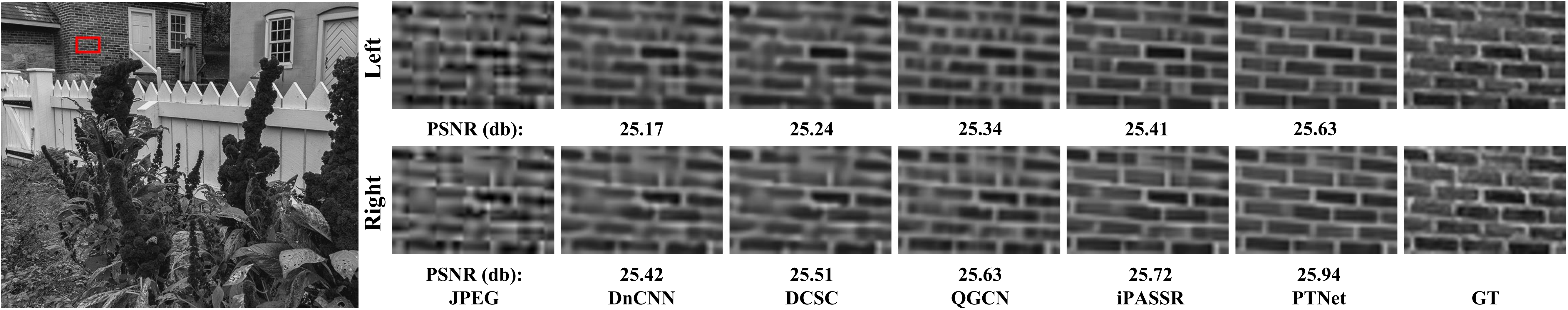}
	\end{center}
	\caption{Visual comparisons on the image ’0095’ from Flickr1024~\cite{wang2019flickr1024} at QF 10. The proposed PTNet is compared with the state-of-the-art methods including DnCNN~\cite{zhang2017beyond}, DCSC~\cite{fu2019jpeg}, QGCN~\cite{li2020learning} and iPASSR~\cite{wang2021symmetric}. Better zoom in.}
	\label{fig:visual_re_2}
\end{figure*}

\begin{figure*}[t]
	\begin{center}
		\includegraphics[width=1\linewidth]{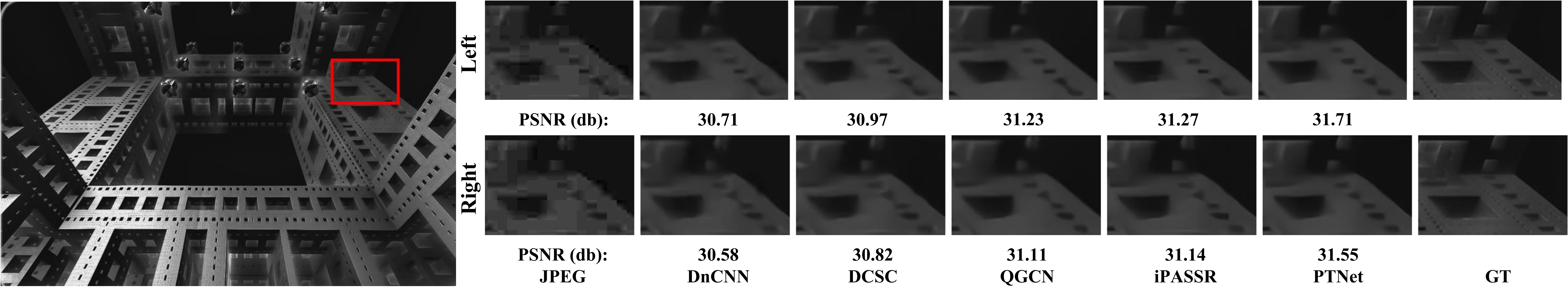}
	\end{center}
	\caption{Visual comparisons on the image ’0082’ from Flickr1024~\cite{wang2019flickr1024} at QF 10. The proposed PTNet is compared with the state-of-the-art methods including DnCNN~\cite{zhang2017beyond}, DCSC~\cite{fu2019jpeg}, QGCN~\cite{li2020learning} and iPASSR~\cite{wang2021symmetric}. Better zoom in.}
	\label{fig:visual_re_4}
\end{figure*}

We provide more qualitative visualization results in Figs.~\ref{fig:visual_re_1},~\ref{fig:visual_re_2} and ~\ref{fig:visual_re_4} to compare our PTNet with other methods. It can be found that the results of our PTNet have better perceptual quality compared to other methods.

\end{document}